\pdfoutput=1
\documentclass{article} 
\usepackage[T1]{fontenc}
\usepackage[preprint]{colm2025_conference}

\usepackage{microtype}
\usepackage{hyperref}
\usepackage{url}
\usepackage{booktabs}

\definecolor{darkblue}{rgb}{0, 0, 0.5}
\hypersetup{colorlinks=true, citecolor=darkblue, linkcolor=darkblue, urlcolor=darkblue}
\usepackage{graphicx}
\usepackage{subcaption}
\usepackage{multirow, multicol}
\usepackage{diagbox}
\usepackage{algorithm}
\usepackage{algpseudocode}
\usepackage{comment}
\usepackage{lineno}

\usepackage{tabularx}
\usepackage{xspace}
\usepackage{amsmath}
\usepackage{float}
\usepackage{xcolor}
\usepackage{tablefootnote}
\usepackage{setspace}

\usepackage{pifont}
\usepackage{amssymb}

\usepackage{enumitem}

\newcommand{\TODO}[1]{{\color{red}TODO: #1}}

\newcommand{\commentGC}[1]{{\color{blue}GC: #1}}

\newcommand{\benchmark}{WebLists\xspace}
\newcommand{\agent}{BardeenAgent\xspace}

\title{\benchmark: Extracting Structured Information From Complex Interactive Websites Using Executable LLM Agents}

\author{Arth Bohra\thanks{Work done while at Bardeen}\\
College of Computing, Data Science, and Society \\
University of California Berkeley \\
Berkeley, CA, USA \\
\texttt{arthbohra@berkeley.edu}\\
\AND
Manvel Saroyan, Danil Melkozerov\textsuperscript{\textdagger}, Vahe Karufanyan\thanks{Equal contribution},\\
{\bf Gabriel Maher, Pascal Weinberger, Artem Harutyunyan, Giovanni Campagna}\\
Bardeen, Inc.\\
San Francisco, CA, USA \\
\texttt{\{manvel,danil,vahe.karufanyan.c,gabriel,pascal,artem,giovanni\}@bardeen.ai}
}

\begin{document}

\ifcolmsubmission
\linenumbers
\fi

\maketitle

\begin{abstract}
Most recent web agent research has focused on navigation and transaction tasks, with little emphasis on extracting structured data at scale. We present \benchmark, a benchmark of 200 data-extraction tasks across four common business and enterprise use-cases. Each task requires an agent to navigate to a webpage, configure it appropriately, and extract complete datasets with well-defined schemas. We show that both LLMs with search capabilities and SOTA web agents struggle with these tasks, with a recall of 3\% and 31\%, respectively, despite higher performance on question-answering tasks.

To address this challenge, we propose \agent, a novel framework that enables web agents to convert their execution into repeatable programs, and replay them at scale across pages with similar structure. \agent is also the first LLM agent to take advantage of the regular structure of HTML. In particular \agent constructs a generalizable CSS selector to capture all relevant items on the page, then fits the operations to extract the data.

On the \benchmark benchmark, \agent achieves 66\% recall overall, more than doubling the performance of SOTA web agents, and reducing cost per output row by 3x.
\end{abstract}

\section{Introduction}{\label{section:introduction}}

Recent progress in the capabilities of Large Language Models has brought increased attention to LLM-based agents that can navigate the web and perform tasks autonomously \citep{gur2022understanding, kagaya2024rap, kim2024language}. As a result, several benchmarks have been proposed to evaluate LLM web agents, both in simulated environments \citep{webshop22, zhou2024webarenarealisticwebenvironment} and on real websites \citep{he2024openwebvoyagerbuildingmultimodalweb, deng2023mind2web}. The results of these benchmarks suggest that state-of-the-art agents with reasoning models can perform comparably to humans \citep{putta2024agentqadvancedreasoning, abuelsaad2024agenteautonomouswebnavigation, agashe2024agentsopenagentic, claudecomputeruse, openaioperator}.

Despite their successes on academic benchmarks,
the adoption of LLM-based web agents in industry has been limited for data extraction and research tasks. 
We identify two limitations of existing evaluations that could explain this phenomenon:
\begin{itemize}
\item Existing benchmarks focus on navigation and form-filling, rather than extracting data. For the few data-extraction tasks, the expected output is unstructured. However, in realistic business use cases agents are combined with traditional programs in a multi-stage pipeline. In this setting, it is necessary to extract structured machine-readable information.


\item Existing benchmarks involve a small number of websites: WebVoyager covers 15 websites \citep{he2024openwebvoyagerbuildingmultimodalweb}, and WebArena only 4 \citep{zhou2024webarenarealisticwebenvironment}. Additionally, there is only one website per task: one recipe website, one code repository, etc., making it easy to inadvertently overfit the agent. In reality, the web has a long tail of websites with diverse markups~\citep{curatelabs}, which differ significantly in layout, interaction patterns, and structure. 

\end{itemize}

\begin{figure*}[t]
  \centering
  \includegraphics[width=\linewidth]{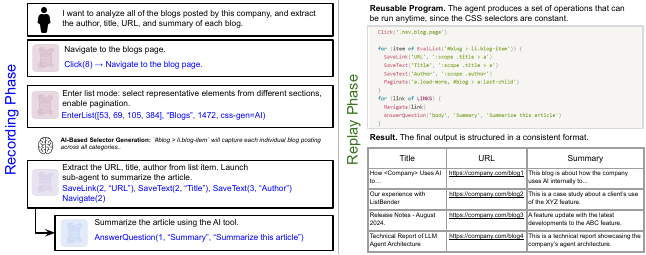}
  \caption{\agent first records all actions it performs on the web page, then converts them into a program with loops, which it executes to extract the data.}
  \label{fig:overview}
  \vspace{-1em}
\end{figure*}

To address these limitations, we propose the problem of interactive schema-bound data extraction across websites: given a high-level task description and a desired output schema, the agent must (1) navigate to the appropriate part of a website, (2) manipulate any relevant interactive elements to expose the target data, and (3) extract it in structured form, while adhering to the schema. 

Despite being highly relevant commercially, this problem is difficult for known methods. On the one hand, the large variety of websites makes traditional information retrieval ineffective. Most websites do not use semantic markup, do not follow good accessibility practices, use obfuscated markups, and use dynamic identifiers.
 Furthermore, the information is often not present in the markup at all until an appropriate form is filled, meaning that live agent interaction with the page is needed to reveal it, and one cannot build an index ahead of time. On the other hand, this problem requires extracting large volumes of data. Existing LLMs can ingest hundreds of thousands or even millions of input tokens \citep{openai2024gpt4technicalreport, geminiteam2024geminifamilyhighlycapable}, but they are limited to a few thousand output tokens. 
Hence, to succeed on this problem, an agent has to break the larger task down into smaller sub-tasks and repeat the sub-tasks over all items to be extracted. Our experiments show that general purpose web agents are not capable of doing so, resulting in low recall: while agents may successfully navigate to the right page, they cannot extract all of the data.

To advance the research in this challenging problem, we propose \benchmark, a set of benchmark tasks across four different use-cases that are representative of real-world product and company research use cases. 

Additionally, we propose a new web agent architecture optimized for structured data extraction, called \agent. Our main insight is that the desired information is usually in list or tabular form on the page, and this fact is reflected in a markup that has a regular structure. If the agent can successfully operate on one item of the list, we can exploit the regularity to generalize to all items. Hence, rather than directly solving the task, we propose that the agent construct a program, on-demand, and use that to solve the task efficiently.

\agent operates in two steps (Fig.~\ref{fig:overview}):
\begin{enumerate}
\item \textit{Record}: the agent navigates and finds the requested information. As it navigates, it records all the actions it performs, using generalizable CSS selectors, which apply on all pages with the same structure. The agent also detects any list structures, and only operates on the first element of any lists. This step terminates when the agent has successfully extracted the first item in the desired dataset.
\item \textit{Replay}: the actions that were recorded by the agent are converted to an executable program; any detected list structure becomes a loop over all items and pages in the list. Running the program then extracts all the desired information.
\end{enumerate}


Our two-phase approach allows the agent to scale to large datasets with fewer LLM calls. Unlike traditional LLM agents, our method does not require more LLM calls as the number of items to extract grows, which eliminates the error compounding effect. Additionally, the computational complexity depends on the depth of the search, not the size of the dataset.

\subsection{Contributions}

The contributions of this paper are as follows:
\begin{itemize}
    \item We introduce a new benchmark, called \benchmark, to evaluate AI agents on interactive data extractions tasks on the web. The benchmark contains 200 tasks across 4 use cases and 50 websites.
    \item We propose a new web agent architecture, called \agent, which leverages the structured nature of HTML. The LLM is applied to a subset of the data items, while generalizable CSS selectors are used to extract the entire dataset.
    \item Experimental results on \benchmark show that \agent is effective, yielding an improvement of 36\% in recall over best existing web agent, while maintaining a high precision of 72\%. \agent also reduces the cost per output row by 3x.
\end{itemize}

\section{The \benchmark Benchmark Task}\label{section:dataset}

The \benchmark benchmark aims to robustly measure whether an agent can extract structured data on a diverse set of websites. In this section, we discuss how we constructed \benchmark and how it differs from other web agent benchmarks.

\subsection{Use Cases}
The \benchmark benchmark originates from real-world data extraction use cases collected from customers of a commercial web automation provider. We picked 3 use cases that are representative of a typical market research workflow (Table~\ref{tab:benchmark_tasks}):
extracting job posting from a company's career page, extracting blog posts and product updates from a company website, and extracting mentions of customers and customer testimonials. These three basic use cases all involve navigation, interaction and data extraction. The data regularity is also varied: job postings are typically structured, while customer testimonials vary substantially in format and presentation even within the same website. Similarly, blog posts can be easily confused with other marketing pages.

In order to make the benchmark more challenging, we also add a variation of the job postings extraction use case where the agent must filter the available jobs for a specific category or location. An agent may use the filtering mechanism in the page, or it can extract all job postings and filter them at a later step.

\begin{table}[tb]
\caption{Overview of use-cases and schemas in the \benchmark benchmark.}
\label{tab:benchmark_tasks}
\centering
\small
\begin{tabular}{>{\raggedright\arraybackslash}p{2.2cm}>{\raggedright\arraybackslash}p{2.8cm}>{\raggedright\arraybackslash}p{5.2cm}r}
\toprule
\textbf{Use-case} & \textbf{Columns} & \textbf{Description} & {\bf \# Rows} \\
\midrule
Blogs & Update Title, Update URL & Extract and summarize blog articles across content hubs. &  2,589\\
\hline
Testimonials & Name, Testimonial, URL & Extract customer testimonials and case-studies. & 2,943\\
\hline
Jobs & Job Title, Job URL & Extract all open job postings. & 5,524\\
\hline
Job Categories & Job Title, Job URL & Filter and extract job postings by category or location. & 1,081\\
\bottomrule
\end{tabular}
\end{table}

\subsection{Benchmark Structure}
For each of the 4 use cases, we identify 50 websites from the list of top 100 cloud companies according to Forbes~\citep{forbestop100}. For each website and use case, we provide a reference solution by manually annotating the URL of the first page containing the data, and then constructing a script that extracts all the data in the desired form. We limit extraction to at most 5 pages of data per task. The number of rows extracted by the reference solution is in Table~\ref{tab:benchmark_tasks}.

The fact that the use-cases involve live websites means the ground-truth data is always changing (e.g. companies adding or removing job postings from their website). As a result, evaluation of the benchmark involves first running the script to extract the most up-to-date dataset, and then running the agent for comparison.

Each use case also defines a desired schema which the agent's output must adhere to. The schema has both a URL column and one or more free-text columns. The URL acts as an identifier of the extracted row and can be compared to the ground-truth data.  The other columns assess whether the agent can extract structured data, regardless of format.


\subsection{Comparison with Existing Benchmarks}

Table~\ref{table:comparison} compares \benchmark against other well-known web agent benchmarks. \benchmark involves a larger number websites, making it more representative of the breadth and complexity of the real world web. The only other benchmark with a comparable number of websites is Mind2Web~\citep{deng2023mind2web}, which uses human demonstrations on recorded websites. Recorded websites are imperfect proxies of real live websites, and human demonstrations become quickly outdated on live websites.

Furthermore, existing benchmarks make use of simulators or LLM-based evaluation to handle the fact that live websites are continually changing. As \benchmark uses website-specific evaluation scripts, it is the first web agent benchmark that can be used with live websites while providing deterministic evaluation metrics that do not deteriorate over time.


\begin{table}
\caption{Comparison of \benchmark with other commonly used web agent benchmarks.}
\label{table:comparison}
\small
\centering
\begin{tabular}{lllrr}
\toprule
{\bf Benchmark} & {\bf Format} & {\bf Eval.} & {\bf \# Websites} & {\bf \# Tasks} \\
\midrule
WebShop \citep{webshop22} & Simulated & Deterministic & 1 & 12,087 \\
Mind2Web \citep{deng2023mind2web} & Recorded & Demonstrations & 137 & 2,350 \\
WebArena \citep{zhou2024webarenarealisticwebenvironment} & Simulated & Deterministic & 4 & 812 \\
WebVoyager \citep{he2024openwebvoyagerbuildingmultimodalweb} & Live & LLM judge & 15 & 400 \\
\hline
\benchmark & Live & Deterministic & 50 & 200 $(50 \times 4)$ \\
\bottomrule
\end{tabular}
\end{table}


\section{Web Agents that Can Capture Structured Information}\label{section:methods}

In this section, we first introduce the general structure of our web agent, then we present how our approach improves large-scale data extraction tasks.

\subsection{Basic Agent Loop}

Following previous work, we model the web agent problem as a Partially Observable Markov Decision Process (POMDP) where the state $s \in \mathcal{S}$ includes the web page and interaction state, the observation $o$ consists of the URL and DOM state, and the task is specified by a goal description $g$ and schema $\mathbf{c} \in C$. At step $t$, the agent performs actions $a = \pi(o_1, \ldots, o_t)$ where the policy model ($\pi$) samples multiple candidates from an LLM before a second LLM selects the best action, similar to \citet{yang2024agentoccamsimplestrongbaseline} and \citet{gu2024llmsecretlyworldmodel}. The action is implemented via tool calling -- see Appendix~\ref{appendix:dsl} for the complete list. Following \citet{lutz2024wilburadaptiveincontextlearning}, a verifier model checks the success of the action, providing feedback to re-tracking when needed, and computing the next steps.

\subsection{LLM-Generated Executable Agents}\label{section:exec-agents}

The basic agent loop is ineffective for data extraction: LLMs have limited output size, and repeated calls to LLMs are prone to compounding errors. To handle large datasets, we introduce the concept of \textit{executable agents}. These agents operate in two phases, a \textit{record} phase and a \textit{replay} phase. We introduce these using an example where the agent navigates to a blog post page and extracts all blog posts (Fig.~\ref{fig:overview}).

\paragraph{Recording Actions} During the record phase, the agent samples and performs actions according to the proposed policy function, adding each action to a comprehensive recording $R$, while simultaneously converting temporary numeric DOM element IDs to robust CSS selectors (e.g., ``Click(\verb|.nav.blog.page|)'') to ensure reliable element identification across page variations. The selector generation process will be detailed in Section~\ref{section:css-generation}.

\paragraph{Recording Lists}

For efficient list extraction, we introduce the EnterList tool, with signature:
$$\text{EnterList}(\textit{elements}, \textit{description}, \textit{pagination})$$
The tool allows the agent to select representative elements, identify their common parent, and create a scoped extraction context, while also specifying pagination controls (Fig.~\ref{fig:overview}, with a detailed example in Fig.~\ref{fig:list-example} in the appendix); after calling this tool, the agent's observation narrows to only the first list item and all CSS selectors become scoped to the list item, enabling focused recording of operations that can be applied to any list item, before returning to normal operation via ExitList.

\paragraph{Replaying Recorded Actions}
After the record phase, \agent creates a content-robust executable program that navigates to the initial page and replays recorded actions. List extraction is handled by looping through CSS-matched elements with proper scoping, filtering out repeated items during pagination to detect the end of the list and prevent infinite loops (especially when ``Next'' buttons become ``Previous'' buttons). \agent also ensures that global actions performed outside lists are properly associated with each extracted entry in the final output table.

\paragraph{Handling Navigation in Lists}
When list items require navigation to access detailed information, the agent avoids immediate navigation, because that would lose the list context. The agent instead extracts all item URLs first, then performs a second pass where it navigates to each collected URL and executes the appropriate actions (Fig.~\ref{fig:overview}).

\subsection{Selector Generation}\label{section:css-generation}


To reliably replay actions across different list items or pages, we propose two methods for generating general and robust CSS selectors: a computationally efficient heuristic approach, and an LLM-based CSS generation model for more complex cases.

\paragraph{Heuristic Generation}

Our heuristic approach samples diverse CSS selector strategies (parent-child, position, HTML classes/IDs, attributes), filters nonmatching candidates, and joins viable selectors. For example, the Next button in Fig.~\ref{fig:overview} is identified as both \verb|a.load-more| (tag name, HTML class) and \verb|#blog > a:last-child| (parent-child, position). For lists, we locate the least common ancestor of selected elements and target its immediate children. 

This multi-strategy approach proves crucial as different selectors excel in different contexts: HTML classes may be semantically rich on some pages but meaningless on others, while positional selectors work well with consistent layouts despite their general brittleness.

\paragraph{LLM-based Generation}
For cases where list elements lack a clean common parent (e.g., blog posts across different categories with headers or dividers intermixed), we propose a novel LLM-based algorithm for CSS selector generation. The agent selects elements from different sections, computes their least common ancestor, and invokes a \textit{selector model}:
$$
s = \text{SelectorModel}(g, \mathbf{c}, H, d)
$$
where $g$ is the goal, $\mathbf{c}$ is the column specification, $H$ is the HTML of the least common parent, and $d$ is the list description. This approach enables generalization beyond immediate element structure to reliably identify all relevant items, even in complex or irregular layouts.

\subsection{Creating Schema-Bound Data Tables}
Due to the downstream need to have structured output for further processing, the agent output must adhere to the column specification $\mathbf{c}$. We accomplish this in three ways:

\paragraph{Direct Extraction}
If the page directly contains the desired data in one specific HTML element or attribute, the agent can extract it directly into the appropriate column without further processing. To maximize applicability, \agent includes all leaf elements in the observation DOM, including formatting elements for single values -- a trade-off that increases LLM input size but preserves the efficiency of direct extraction.



\paragraph{Regular Expression}
In some cases, the desired data might be identifiable via a regular expression (e.g., emails or prices), prompting the agent to compute a matching pattern when using the SaveText tool and apply it during replay. This approach maintains the efficiency of direct extraction while producing clean, machine-readable output.


\paragraph{Question Answering}
If no element exclusively contains the desired data and no regular expression can extract it, the agent employs a special AnswerQuestion tool that uses an LLM with a question-answering prompt on the extracted text. During replay, this same prompt processes each newly extracted element. This tool the only instance where an LLM is applied to multiple list elements, and serves as a fallback for structured output when page structure doesn't match user specifications, without affecting control flow decisions and thus avoiding compounding errors.




\section{Evaluation}\label{section:experiments}

In this section, we evaluate the state-of-the-art methods on \benchmark, and also show that our proposed \agent improves over their performance.

\subsection{Models}
We evaluate the following methods:

\begin{itemize}
\item \textit{Wilbur}~\citep{lutz2024wilburadaptiveincontextlearning} is web agent that employs tool calling, self-verification and backtracking, optimized for the WebVoyager benchmark~\citep{he2024openwebvoyagerbuildingmultimodalweb}. 
\item \textit{Agent-E}~\citep{abuelsaad2024agenteautonomouswebnavigation} is a web agent that combines a planner agent and a browser control agent. The latter uses tool calling to observe the browser state and perform actions. At the time of writing, Agent-E is the state-of-the-art open-source agent for WebVoyager.
\item \textit{LLM + Search}: as a non-agentic comparison, we evaluate a commercial LLM system that indexes the web ahead of time, retrieves data from the index according to the stated goal, and then uses the retrieved data to answer the question. We use Perplexity~\citep{perplexity} for this purpose.
\end{itemize}

We also experimented with other commercial systems. It is too expensive for us to evaluate them fully on \benchmark, but we report results with OpenAI Operator~\citep{openaioperator} on the Jobs use-case in Appendix~\ref{appendix:operator-eval}.



\subsection{Experimental Setup}

\paragraph{Hyperparameters}

We tune goals per use-case and model (without website-specific optimization) and use manually selected few-shot examples for \agent and Wilbur, rather than Wilbur's noisy auto-curriculum. For Agent-E and LLM+Search, we construct goals requesting Markdown table outputs. Additional hyperparameter details are in Appendix~\ref{appendix:hparams}.


\paragraph{Metrics}

We evaluate precision (percentage of agent-extracted items appearing in the gold set) and recall (percentage of gold set items extracted by the agent), with high precision identifying potential hallucinations and high recall ensuring complete dataset extraction. We average these metrics across websites, using exact matching on a specific column to compare gold and retrieved data\footnote{For URL columns, we disregard any query parameter, as they often contain unique randomly generated tracking IDs.}. We choose different columns for different agents, in order to maximize matching between gold and extracted data.






\subsection{Results}

Results are shown in Table~\ref{table:results}. On average across all four use cases, we see that \agent's recall is 36\% higher than the second best method. Both Wilbur and Agent-E extract pages one by one, and due to compounding errors in the policy and planner/verifier, they are prone to terminating the execution too early, resulting in low recall. \agent's substantially higher recall stems from its ability to paginate and scrape large amounts of data at once.

The average precision of Wilbur is 10\% higher than \agent. We attribute this to the fact that entering list mode is a critical step in \agent, and it is prone to errors (e.g., selecting wrong elements or over-generalizing CSS selectors) which can result in additional irrelevant elements. This is especially evident in the Job Categories use case, where incorrect page filtering can lead to extracting jobs across all categories instead of just the targeted ones. In practice, an additional LLM filtering step could be applied to the output data to remove irrelevant rows.

Finally, we note that the baseline LLM with search has very low recall. We hypothesize this is because the retrieval from the index is unsuitable for structured data. On Job Categories specifically, the recall is 0.2\%, as this data only appears after interaction with filters and is therefore absent from the index. The LLM with search also suffers from low precision due to model hallucinations and outdated indexed data.


\begin{table}[tbp]
\centering
\caption{Evaluation metrics across use-cases in the \benchmark benchmark.}
\label{table:results}
\small
\begin{tabular}{lrrrrrrrrrr|r}
\toprule
 & \multicolumn{2}{c}{\textbf{Jobs}} 
 & \multicolumn{2}{c}{\textbf{Job Cat.}} 
 & \multicolumn{2}{c}{\textbf{Blogs}} 
 & \multicolumn{2}{c}{\textbf{Testimonials}} 
 & \multicolumn{2}{c|}{\bf Overall}
 & \textbf{Q\&A} \\
 
\textbf{Agent} 
 & \textbf{Pre} & \textbf{Rec} 
 & \textbf{Pre} & \textbf{Rec} 
 & \textbf{Pre} & \textbf{Rec} 
 & \textbf{Pre} & \textbf{Rec} 
 & \textbf{Pre} & \textbf{Rec} 
 & \textbf{Acc.} \\
\midrule
\text{Agent-E} &  50.0 & 17.3 & 22.0 & 9.0 & 36.7 & 8.7 & \bf 68.5 & 13.5 & 44.3 & 12.1 & 58.0 \\
\text{Wilbur} & \bf 95.2 & 18.7 & \bf 86.9 & 62.5 & \bf 90.4 & 24.6 & 55.4 & 16.1 & \bf 82.0 & 30.5 & 56.0 \\
\text{LLM + Search} & 23.3 & 2.1 & 13.3 & 0.2 & 27.9 & 8.4 & 13.7 & 2.6 & 19.6 & 3.3 & 42.0 \\
\hline
\text{\agent} & 87.0 & \bf 86.8 & 71.1 & \bf 64.5 & 71.9 & \bf 59.9 & 59.9 & \bf 53.5 & 72.5 & \bf 66.2 & \bf 60.0 \\
\text{$-$ Selector Model} & 78.5 & 25.0 & 63.9 & 45.7 & 81.8 & 32.7 &  59.2 & 38.8 & 70.9 & 35.6 & \bf 60.0 \\
\bottomrule
\end{tabular}
\end{table}

\subsection{Cost Evaluation}\label{appendix:cost-eval}

In Table~\ref{table:cost-results} we show the number of tokens and the cost used by the four agentic methods, averaged across all 200 tasks. We also report the total number of correctly extracted rows, and the amortized cost per row. We can see that baseline methods have lower cost per website due to the reduced complexity of the agent, with Agent-E being the most efficient with a highly optimized DOM representation. At the same time, baseline methods retrieve substantially fewer rows, and consume more (expensive) output tokens than input tokens, so their amortized cost is more than 3x than that of \agent.

\begin{table}[tb]
\caption{Number of tokens and cost (in cents USD), both average per task (website, use-case pair) and average per correct output row.}
\label{table:cost-results}

\centering
\begin{tabular}{lrrrrrrr}
\toprule
& \multicolumn{3}{c}{\bf Per-task} & & \multicolumn{3}{c}{\bf Per-row} \\
{\bf Agent} & {\bf Input} & {\bf Output} & {\bf Cost} & {\bf \# Rows} & {\bf Input} & {\bf Output} & {\bf Cost} \\
\midrule
Agent-E & \bf 23,436 & \bf 1,155 & \bf 26.9 & 1,675 & \bf 2,798 & 138 & 3.21 \\
Wilbur &  108,294 & 1,594 & 28.9 & 1,269 & 17,067 & 251 & 4.55 \\
\agent & 192,695 & 1,624 & 44.4 & \bf 8,269 & 4,660 & \bf 39 & \bf 1.07 \\
$-$ Selector Model & 157,072 & 1,418 & 41.0 & 3,061 & 10,263 & 93 & 2.68\\
\bottomrule
\end{tabular}
\end{table}

\subsection{Question-Answering Evaluation}

In addition to structured data extraction, we evaluate our approach on traditional question-answering to demonstrate it maintains generality. We collected 50 questions (one per website in \benchmark), comprising product and company-related questions ranging from simple retrieval to complex calculations that require reasoning about requirements and operating ROI or cost calculator widgets. Examples of these questions are in Appendix~\ref{appendix:qa-examples}.

Results are shown in Table~\ref{table:results}. We report the accuracy, measured by manual evaluation, against a human annotated answer. Methods optimized for question answering, such as the LLM with search, perform reasonably well with 42\% accuracy, despite having around 3\% recall on \benchmark. Yet, all agentic methods, outperform the search LLM by at least 14\%. We observed that this was due to factors like the search LLM not being able to set up a page before answering a question or having outdated indexes. \agent performs similar to the other state-of-the-art methods, demonstrating that our executable approach maintains strong performance even for tasks that don't involve list extraction.

\subsection{Discussion}\label{section:discussion}

\paragraph{Balacing Extraction \& Interaction}

\agent's primary advantage is its ability to scrape lists and paginate systematically, and its prompts are optimized for data extraction. This can come at a cost, in terms of the ability to operate complex forms and widgets. For example, the Job Categories use case proved challenging as it often requires complex setup procedures involving multiple filters and selections before the relevant data becomes accessible. Future work should investigate specialized tools to interact with complex widgets, or different DOM representations for different stages of the task.

\paragraph{Selecting The Right List Elements}

For \agent, precision sometimes suffers due to overly inclusive lists. In some cases, this is due to definitional ambiguities. For instance, blog pages often contain various types of links (promotional, related posts, etc.) that may be incorrectly interpreted as primary content.

Performance differences are most pronounced with multi-category data extraction. For example, for job postings spanning various departments, heuristic CSS selectors require extracting each category as a separate list, increasing latency and cost. LLM-based selector generation efficiently extracts across categories simultaneously. However, for simpler single-category use cases (e.g., blogs), the agent without the selector model generation delivers higher precision. 
Future work should focus on improving the reliability of the selector model and exploring whether multimodal models could better determine appropriate selectors based on website layout, potentially balancing precision and recall more effectively.

\paragraph{Getting to All Pages}
For blog posts with numerous pages, \agent's pagination capabilities shine compared to competitors that must individually navigate to each post, significantly reducing token usage and cost. Attempting to fix this issue for Agent-E and Wilbur with prompts such as "visit every page" result in higher cost compared to our executable agent approach, and still fail to extract the whole dataset, because each page requires individual navigation and separate LLM invocations, which leads to compounding errors.

\paragraph{Extracting Vs. Generating}

Rule-based extraction shows significant advantages over question-answering approaches. While we have evaluated with a forgiving metric, requiring only one column to match, a stricter evaluation would yield lower recall. In particular, Agent-E could not extract links at all, due to its highly optimized DOM representation which nonetheless omits critical information. 

Conversely, we found \agent can sometimes fail with question-answering tasks, by extracting text that merely contains the answer, rather than providing the desired precise answer. The AnswerQuestion tool, while powerful for extracting unstructured data, faces limitations due to the limited context given, sometimes resulting in "I don't know" responses rather than complete answers. We found this to be especially prevalent on yes-no questions with a correct `no' answer, as the agent would assume that the data was there but it just couldn't find it, and respond with `I don't know'. An agent optimized for question-answering, such as Agent-E, outperforms \agent on these types of questions. Future work could look into balancing the trade-off between efficient list-based extraction and highly specific data capture.

\section{Related Works}\label{section:background}

\paragraph{Web Agents}

A growing body of work has explored the use of LLMs to create general-purpose agents that can navigate the Web \citep{lutz2024wilburadaptiveincontextlearning}. Web agents are a promising technology for personalized automated web navigation due to the ability of LLMs to handle user-provided instructions \citep{cai2024largelanguagemodelsempowered}. However, more work is needed to further improve the robustness of LLM-based web agents. The performance of LLM web agents has been shown to be sensitive to the choice of state and action space \citep{yang2024agentoccamsimplestrongbaseline}. To this end, planning algorithms have been explored as a way to improve LLM-based web agents \citep{putta2024agentqadvancedreasoning, gu2024llmsecretlyworldmodel, he2024openwebvoyagerbuildingmultimodalweb, qi2025webrltrainingllmweb}. Robust handling of interactive and dynamic web elements, such as forms, remains a challenge for web agents \citep{pan2024webcanvasbenchmarkingwebagents, thomas2025webgameschallenginggeneralpurposewebbrowsing, agashe2024agentsopenagentic, abuelsaad2024agenteautonomouswebnavigation, xu2025turkingbenchchallengebenchmarkweb, song2025bearcubsbenchmarkcomputerusingweb}.

\paragraph{Structured Data Extraction}

Automated data extraction from web pages has long been recognized as an important technology.
As such an abundance of methodologies and software tools have been developed to enable automated scraping of web pages \citep{bishit23, dikilita24}.
Several key challenges in its implementation are web page navigation, locating relevant information on the page and extracting data in a structured form \citep{barbaresi-2021-trafilatura, bevendorff23, wahed24}. Complications also arise from the fact that real web pages typically contain a plethora of elements other than the data to be extracted \citep{yi03}.
One way of managing these challenges is to focus on specific categories of data, e.g. news articles \citep{dallabetta2024fundussimpletousenewsscraper, bedrin2025multilingualattributeextractionnews, SHAH2025110284}. Other methods have focused on developing models of the common structure of web pages \citep{vogels18, jung22, JUNG2023101501}.
As webpages are based on HTML and the DOM, their structure has also been exploited to create web scrapers \citep{Sun2011DOMBC}.
Despite the utility of web scraping and abundance of available tools and methodologies, there is a lack of criteria and datasets with which to evaluate the performance of web scraping methods.

\section{Conclusion}\label{section:conclusion}

In this paper, we have presented \benchmark, the first benchmark focused on large-scale data extraction that requires agents to navigate to the right page, interact with it to reveal the data, and then extract it while adhering to a desired schema. Our experiments show that existing methods perform poorly on this benchmark, with a recall of at most 30\%.

To address this challenge, we also propose \agent, a novel method to scale web agents to large datasets. The core idea is that rather than solving the task directly, the agent should construct an efficient program that solves the task. This is accomplished by recording one instance and generalizing it via CSS selectors that leverage the structured nature of HTML. \agent improves the recall on \benchmark by 36\% (double the previous best result), while maintaining over 72\% precision. We also show that, with a 60\% accuracy, \agent is effective on more traditional question-answering tasks, indicating that generalizing to large scale dataset need not come at the expense of fine-grained complex questions.

While the results are promising, real-world business use cases require higher performance still, especially once datasets are scaled further, and agents are applied to more sensitive use cases. We hope that by releasing the benchmark, we can encourage progress on this important challenge.




\bibliography{custom, acl_anthology}
\bibliographystyle{colm2025_conference}

\newpage

\appendix
\section{Agent DSL}\label{appendix:dsl}

The \agent agent uses a DSL with the following available functions:

{
\centering

\begin{tabular}{lp{6.5cm}}
\toprule
{\bf Name} & {\bf Description} \\
\midrule
Click$(\textit{element})$ & Click on an element in the page \\
\hline
TypeInput$(\textit{element}, \textit{value}, \textit{enter})$ & Enter the given value in a text box or input element. If \textit{enter} is \textsc{True}, emulate pressing the Enter key on the keyboard afterwards. \\
\hline
GoBack$()$ & Navigate to the previous URL in the agent history. \\
\hline
EnterList$(\textit{elements}, \textit{description}, \textit{pagination})$ & Begin capturing a list. \\
\hline
ExitList$()$ & Complete capturing the list. \\
\hline
SaveCurrentURL$(\textit{column})$ & Save the URL of the current page to the given output column. \\
\hline
SaveText$(\textit{element}, \textit{column}, \textit{regex})$ & Save the text content of the given element to the given column. If \textit{regex} is specified, only save the portion of the text that matches the regular expression. \\
\hline
SaveLink$(\textit{element}, \textit{column})$ & Save the link target of the given element to the given column. \\
\hline
AnswerQuestion$(\textit{element}, \textit{column}, \textit{question})$ & Use an LLM to answer the question given the text content of the element, then save the answer to the given column. \\
\hline
Finish$()$ & End execution and complete the task. \\
\hline
Fail$(\textit{errorCode}, \textit{errorMessage})$ & Abort execution due to an error detected by the agent, such as CAPTCHA, 404, network errors, etc. \\ 
\bottomrule
\end{tabular}
}

\section{List Extraction Example}

\begin{figure}[htb]
\small
\textbf{Original DOM}
{
\footnotesize
\begin{tabbing}
01\=23\=45\=67\=\kill
\verb|<div id="blog">|\\
\>\verb|<ul>|\\
\>\>\verb|<li class="blog-item">|\\
\>\>\>\verb|<div>|\\
\>\>\>\>\verb|<div class="title"><a href="/new-release">New Release: v3.14.15</a></div>|\\
\>\>\>\>\verb|<div class="author">John P. Emm</div>|\\
\>\>\>\verb|</div>|\\
\>\>\verb|</li>|\\
\>\>\verb|<li class="blog-item">|\\
\>\>\>\verb|<div>|\\
\>\>\>\>\verb|<div class="title"><a href="/pricing">Pricing updates for 2025</a></div>|\\
\>\>\>\>\verb|<div class="author">Jane Salesperson</div>|\\
\>\>\>\verb|</div>|\\
\>\>\verb|</li>|\\
\>\>...\\
\>\verb|</ul>|\\
\>\verb|<a class="load-more" href="/list?page=2">Next page</a>|\\
\verb|</div>|
\end{tabbing}
}
\textbf{Simplified DOM}
{
\footnotesize
\begin{tabbing}
01\=23\=45\=67\=\kill
\verb|<ul #1>|\\
\>\verb|<li #2>|\\
\>\>\verb|<a #3 href="/new-release">New Release: v3.14.15</a>|\\
\>\>\verb|<div #4>John P. Emm</div>|\\
\>\verb|</li>|\\
\>\verb|<li #5>|\\
\>\>\verb|<a #6 href="/pricing">Pricing updates for 2025</a>|\\
\>\>\verb|<div #7 class="author">Jane Salesperson</div>|\\
\>\verb|</li>|\\
\>\>...\\
\verb|</ul>|\\
\verb|<a #8 class="load-more" href="/list?page=2">Next page</a>|\\
\end{tabbing}
}
\begin{minipage}{0.44\linewidth}
\textbf{Policy model}\\
\setstretch{1.1}
$\text{EnterList}([3, 6], \text{``Blog posts''}, 8)$\\
$\text{SaveText}(3, \text{``Update Title''})$\\
$\text{SaveLink}(3, \text{``Update URL''})$\\
$\text{SaveText}(4, \text{``Author''})$\\
$\text{ExitList}()$
\end{minipage}
\quad
\begin{minipage}{0.55\linewidth}
\textbf{Generated CSS}
\begin{tabbing}
123\=\kill
List: \verb|#blog li.blog-item|\\
Update Title: \verb|:scope .title > a|\\
Update URL: \verb|:scope .title > a|\\
Author: \verb|:scope .author|\\
Pagination: \verb|a.load-more, #blog > a:last-child|
\end{tabbing}
\end{minipage}
\caption{Example of \agent extracting data from a blog, with pagination. The policy model sees a simplified DOM with unnecessary elements and attributes removed, and refers to each element via numerical ID. The IDs are then converted to CSS selectors.}
\label{fig:list-example}
\end{figure}

\section{Experimental Details}\label{appendix:hparams}

The policy model of \agent uses GPT 4 Turbo with temperature 0.7 to sample the initial candidate actions, and GPT 4o with temperature 0.2 to choose from this list. The verifier uses GPT 4 Turbo with temperature 0. The question-answering model for the AnswerQuestion tool uses Gemini Flash 2.0 with temperature 0. The selector model uses Gemini 1.5 Pro with temperature 0.4. Nucleus sampling is not applied. All models are text-only. All models use chain-of-thought prompting.

 For Agent-E and Wilbur we use the same hyperparameters as in their respective papers; the main model used in those agents is GPT 4 Turbo. We use Sonar~\citep{perplexity} as the LLM with search. 

\section{Evaluation of OpenAI Operator}\label{appendix:operator-eval}

Operator is a general purpose web agent recently released by OpenAI.
We evaluated Operator on the Jobs use case of \benchmark, using the same prompt as was used for the LLM + Search evaluation. 
Evaluation on more benchmarks was hindered by the fact that Operator is available only in experimental chat mode and, at the time of this writing, does not provide a programmatic API.

Similar to our other comparisons we measured the precision and recall of job postings retrieved by Operator (Table~\ref{tab:operator:performance_metrics}). 
We find that Operator typically has poor recall, often extracting only a subset of job postings despite being asked to extract all of them.
Indeed on 20 of the 50 companies Operator did not return any job posting at all.
Compared to its recall, Operator has better precision, but it is still substantially below that of \agent.

\begin{figure}
\centering
\includegraphics[scale=0.55]{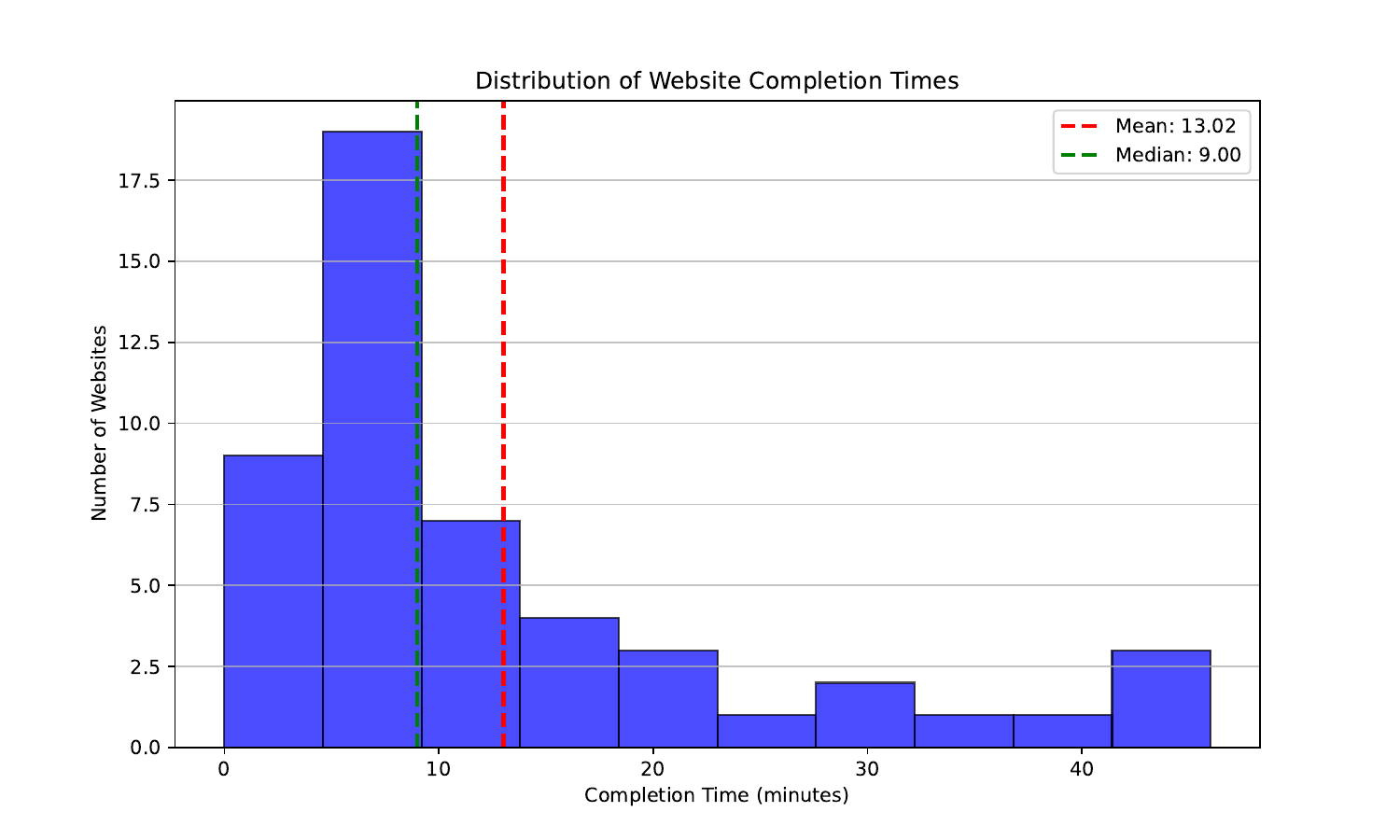}
\caption{Completion times for OpenAI Operator agent on the \benchmark Jobs use case.}
\label{fig:operator:time}
\end{figure}

In Fig.~\ref{fig:operator:time} We also show the the time for Operator to complete a request end-to-end. The median time is 9 minutes per website, much higher than other agents. Operator's completion time is highly variable, and takes longer than 30 minutes on many occasions.

\begin{table}[tbp]
    \centering
    \caption{Performance metrics on Jobs benchmark for OpenAI Operator agent.}
    \begin{tabular}{lc}
        \toprule
        \textbf{Metric} & \textbf{Value} \\
        \midrule
        Precision & 0.41 \\
        Recall & 0.08 \\
        \bottomrule
    \end{tabular}
    \label{tab:operator:performance_metrics}
\end{table}

\section{Question Answering Examples}\label{appendix:qa-examples}

\begin{table}[htb]
\caption{}
\small
\centering
\begin{tabular}{lp{11cm}}
\toprule
{\bf Website} & {\bf Question} \\
\midrule
databricks.com & I am thinking about switching to Databricks for my AI development needs and want to calculate the costs. How much would I be paying per month for the following compute type: All-Purpose compute, m4.xlarge | 4 CPUs | 16GB AWS instance type, with 10 active instances that run 20 hours/day for 25 days per month. \\
\hline
openai.com & I am building a web agent that requires several different OpenAI models in it's loop. There's an planner, executor, and a verifier that are all being used to complete a task. Each of these components uses a different model and on average, a different number of tokens. Currently, we use o1 for the executor (we average around 200 million input tokens and 3 million output tokens per month), we use GPT 4.5 for the planner (average 5 million input tokens per month, 1 million output tokens per month), and GPT-4o mini for our verifier (135 million input tokens and 2 million output tokens on average per month). Since API costs are getting far cheaper, assuming other costs (like deployment) are negligible, and that my web agent produces \$4,000 in revenue per month, how profitable is my web agent. Round to the nearest cent. \\
\hline
canva.com & What is my Projected ROI if my country of operation is the US, I'm in real estate, our ARR is between 10m-50m, we have 30 employees, with around 20 expected canvas users, with often use of workplace tools, and we use Adobe Creative Cloud, Confluence, Google Workspace, Notion, and 10k estimate spend on outsourced marketing? \\
\hline
servicetitan.com & My company is called [...] and we are in the chimney sweeping industry. Calculate our ROI if we have 10 techs, 20 office staff, \$700-800 average ticket, 20 average calls, 60-70 percent calls booked, 80-90\% leads to estimates rate, 80-70\% estimates to sales rate, and we are not a customer of Service Titan yet. \\
\hline
deel.com & I want to compare Belgium and Netherland's employment statistics using Deel's employment comparator tool. I am hiring my next employee from one of those countries based on the minimum monthly salary average in those two countries according to the tool. Which country has a lower average minimum monthly salary average and by how much. Round to the nearest 100 Euros. \\
\bottomrule
\end{tabular}
\end{table}

\end{document}